%% file: main.tex
\documentclass[10pt,twocolumn,letterpaper]{article}

\usepackage[pagenumbers]{cvpr} 

\input{preamble}

\definecolor{cvprblue}{rgb}{0.21,0.49,0.74}
\usepackage[pagebackref,breaklinks,colorlinks,allcolors=cvprblue]{hyperref}

\def\paperID{*****} 
\def\confName{CVPR}
\def\confYear{2026}

\title{Generalizable Knowledge Distillation from Vision Foundation Models for Semantic Segmentation}

\author{Chonghua Lv\textsuperscript{1} \textsuperscript{*}, Dong Zhao\textsuperscript{2} \textsuperscript{*}, Shuang Wang\textsuperscript{1} \textsuperscript{\Letter}, Dou Quan\textsuperscript{1}, Ning Huyan\textsuperscript{3}, Nicu Sebe\textsuperscript{2}, Zhun Zhong\textsuperscript{4} \textsuperscript{\Letter}\\
\textsuperscript{1}School of Artificial Intelligence, Xidian University, China\\
\textsuperscript{2}Department of Information Engineering and Computer Science, University of Trento, Italy\\
\textsuperscript{3}Department of Automation, Tsinghua University, China \\
\textsuperscript{4}School of Computer Science and Information Engineering, Hefei University of Technology, China \\
}

\begin{document}
\maketitle
\begingroup
\renewcommand{\thefootnote}{*}
\footnotetext{Equal contribution.}
\endgroup
\begingroup
\renewcommand{\thefootnote}{\Letter}
\footnotetext{Corresponding author.}
\endgroup
\input{sec/0_abstract}
\input{sec/1_intro}
\input{sec/2_related}
\input{sec/3_methods}
\input{sec/4_exps}
\input{sec/5_conclusion}
{
    \small
    \bibliographystyle{ieeenat_fullname}
    \bibliography{main}
}


\end{document}

%% file: preamble.tex
\usepackage{algorithm}
\usepackage{algorithmic}
\usepackage{graphicx}
\usepackage{booktabs}
\usepackage{makecell}
\usepackage{multirow}
\usepackage{colortbl}
\usepackage{amssymb}
\usepackage{amsmath}
\usepackage{marvosym}

\usepackage{bbding}
\usepackage{subcaption}



%% file: sec/0_abstract.tex
\begin{abstract}
Knowledge distillation (KD) has been widely applied in semantic segmentation to compress large models, but conventional approaches primarily preserve in-domain accuracy while neglecting out-of-domain generalization, which is essential under distribution shifts. This limitation becomes more severe with the emergence of vision foundation models (VFMs): although VFMs exhibit strong robustness on unseen data, distilling them with conventional KD often compromises this ability. We propose Generalizable Knowledge Distillation (GKD), a multi-stage framework that explicitly enhances generalization. GKD decouples representation learning from task learning. In the first stage, the student acquires domain-agnostic representations through selective feature distillation, and in the second stage, these representations are frozen for task adaptation, thereby mitigating overfitting to visible domains. To further support transfer, we introduce a query-based soft distillation mechanism, where student features act as queries to teacher representations to selectively retrieve transferable spatial knowledge from VFMs. Extensive experiments on five domain generalization benchmarks demonstrate that GKD consistently outperforms existing KD methods, achieving average gains of +1.9\% in foundation-to-foundation (F2F) and +10.6\% in foundation-to-local (F2L) distillation. The code will be available at \textcolor{blue}{https://github.com/Younger-hua/GKD}.
\end{abstract}

%% file: sec/1_intro.tex
\section{Introduction}
\begin{figure}[!t]
	\centering
	\includegraphics[width=0.99\linewidth]{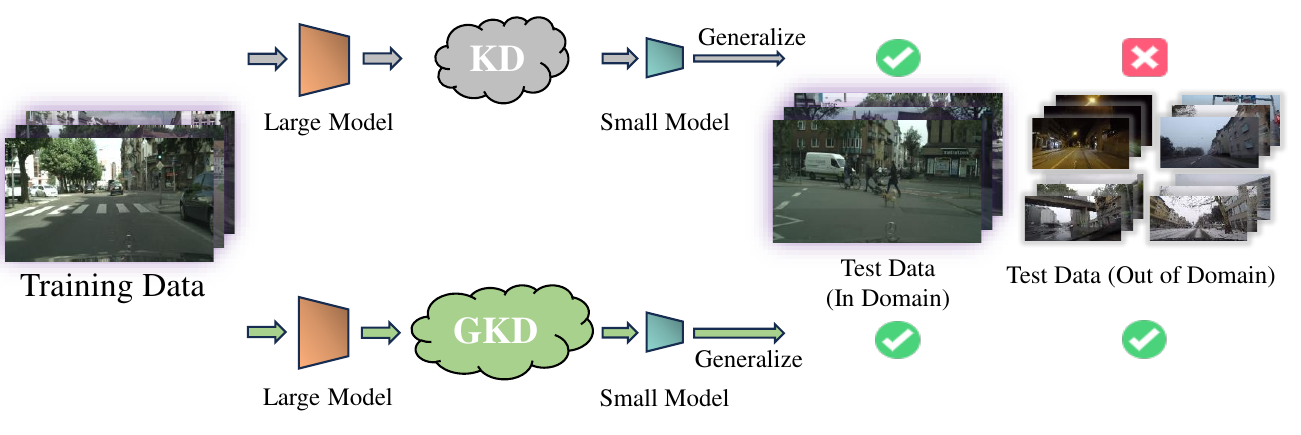}
	\setlength{\abovecaptionskip}{-0.1 cm}
	\caption{Comparison of Knowledge Distillation (KD) and our proposed generalizable KD (GKD). Conventional KD preserves accuracy within the same domain but overlooks generalization to unseen domains.}
	\label{fig_GKD}
	\vspace{-0.6 cm}
\end{figure}

Knowledge distillation (KD) is widely used to compress high-capacity networks into lightweight deployable models for semantic segmentation, reducing the heavy computational and memory cost of dense prediction~\cite{yang2022cross,liu2024transkd,Yang_2024_CVPR,huang2025distilling}. Most KD methods prioritize preserving in-domain accuracy after compression, while paying little attention to domain generalization (DG), as illustrated in \cref{fig_GKD}. DG is particularly critical for segmentation due to frequent domain shifts. For example, autonomous driving systems must generalize across diverse weather and lighting conditions, while medical image segmentation models encounter distribution shifts across devices and clinical sites~\cite{kang2022style,termohlen2023re,zhao2022style}.

This limitation becomes even more pronounced with the emergence of vision foundation models (VFMs)~\cite{oquab2023dinov2,radford2021clip,fang2024eva}, which are widely adopted as universal feature extractors combined with lightweight decoders~\cite{wei2024stronger, Zhao_2025_CVPR}. While VFMs exhibit strong generalization on unseen domains, distilling from them into smaller models via conventional KD often fails to transfer this generalization ability, thereby magnifying the generalization bottleneck. 
A natural question thus arises: \textbf{Can we distill VFMs into compact models to reduce computational overhead without sacrificing their out-of-domain generalization?} In this context, the generalization ability of the distilled model is at least as important as its in-domain accuracy.

\begin{figure}[!t]
	\centering
	\includegraphics[width=0.99\linewidth]{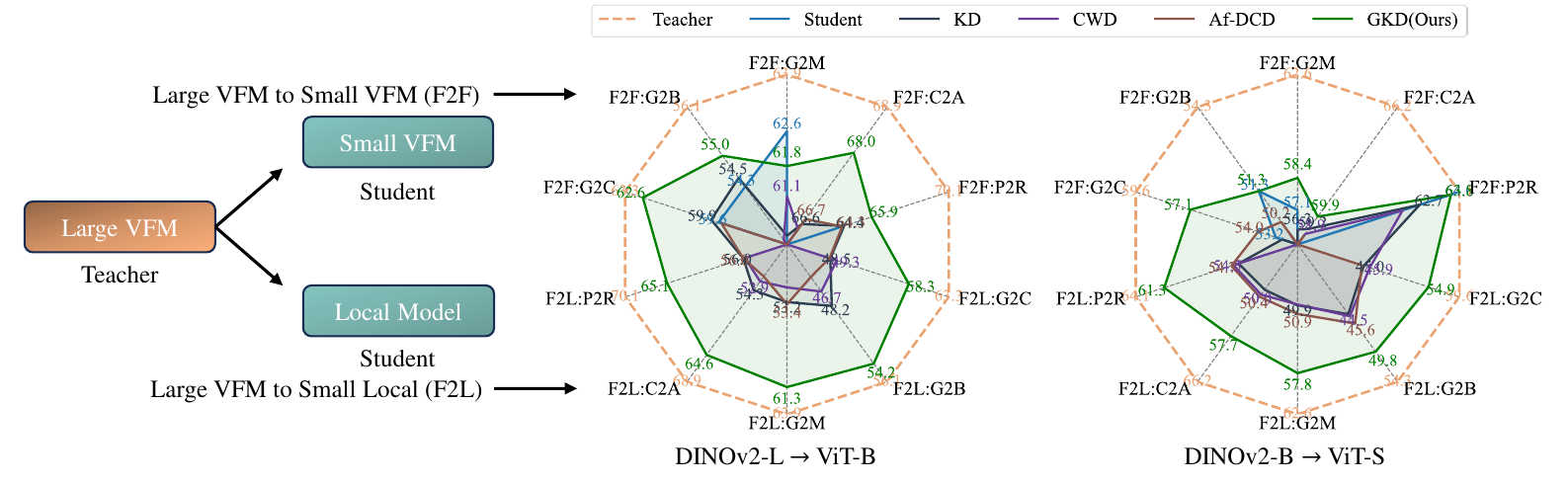}
	\setlength{\abovecaptionskip}{-0.1 cm}
	\caption{Generalization comparison of KD, its enhanced variants (CWD, Af-DCD), and our GKD. GKD consistently outperforms existing KD methods on unseen domains.}
	\label{fig_perf_GKD}
	\vspace{-0.6cm}
\end{figure}

To systematically evaluate the generalization of KD, we consider two representative settings (\cref{fig_perf_GKD}): foundation-to-foundation (F2F), where both teacher and student are VFMs (e.g., DINOv2-L $\to$ DINOv2-B), and foundation-to-local (F2L), where the teacher is a large VFM and the student is a small locally trained model (e.g., DINOv2-B $\to$ ViT-S). Surprisingly, our empirical results reveal that conventional KD often fails to enhance, and can even harm the generalization ability of students.
As illustrated in \cref{fig_perf_GKD}, traditional feature-based KD methods, as well as their enhanced variants (CWD, Af-DCD), consistently produce students that generalize worse than their teachers across unseen domains.
This effect is particularly pronounced in the F2L setting, where the student inherently suffers from weaker generalization. Instead of mitigating domain overfitting, these methods transfer teacher biases tied to the visible domains, amplifying the gap between in-domain and out-of-domain performance. These findings highlight a critical limitation: conventional KD compresses capacity but compromises robustness, underscoring the need for a new distillation paradigm tailored for out-of-domain generalization.

Motivated by these observations, we design a new distillation paradigm that fundamentally departs from the conventional ``single-stage'' KD practice. Our key insight is that representation learning and task learning should not be entangled. We adopt a multi-stage distillation strategy that first extracts domain-agnostic knowledge and only later adapts to the supervised task. Concretely, in the first stage, the student learns generalizable representations through selective feature distillation, while in the second stage, these representations are frozen and leveraged for downstream task learning. This phased design ensures that the student internalizes transferable knowledge before specialization, thereby mitigating domain overfitting and improving cross-domain generalization.

To realize selective feature distillation, we introduce a query-based soft distillation mechanism, where student features act as queries to selectively retrieve spatial knowledge from the teacher via attention. 
This design leverages the rich spatial structure encoded by VFMs, thereby enabling the student to capture only those aspects of the teacher’s knowledge that generalize beyond the visible domain. By decoupling the learning phases and equipping distillation with a query-based mechanism, our framework transforms KD from mere compression into a tool for robust generalization.

Our contributions are three-fold: (1) we empirically diagnose the generalization bottleneck of conventional KD in segmentation; (2) we propose GKD, a new paradigm that decouples representation and task learning through multi-stage distillation and introduces a query-based soft mechanism tailored for VFMs; and (3) we validate GKD on five domain generalization benchmarks under both F2F and F2L settings, where GKD achieves consistent gains of +1.9\% in F2F and a remarkable +10.6\% in F2L, establishing a new state of the art in generalizable distillation. Notably, GKD yields substantial advantages in the label-scarce F2L setting, significantly enhancing label efficiency while maintaining robust cross-domain generalization.

%% file: sec/2_related.tex
\section{Related work}

\begin{figure*}[!tbp]
	\centering
	\begin{subfigure}[b]{0.48\linewidth}
		\centering
		\includegraphics[width=\linewidth]{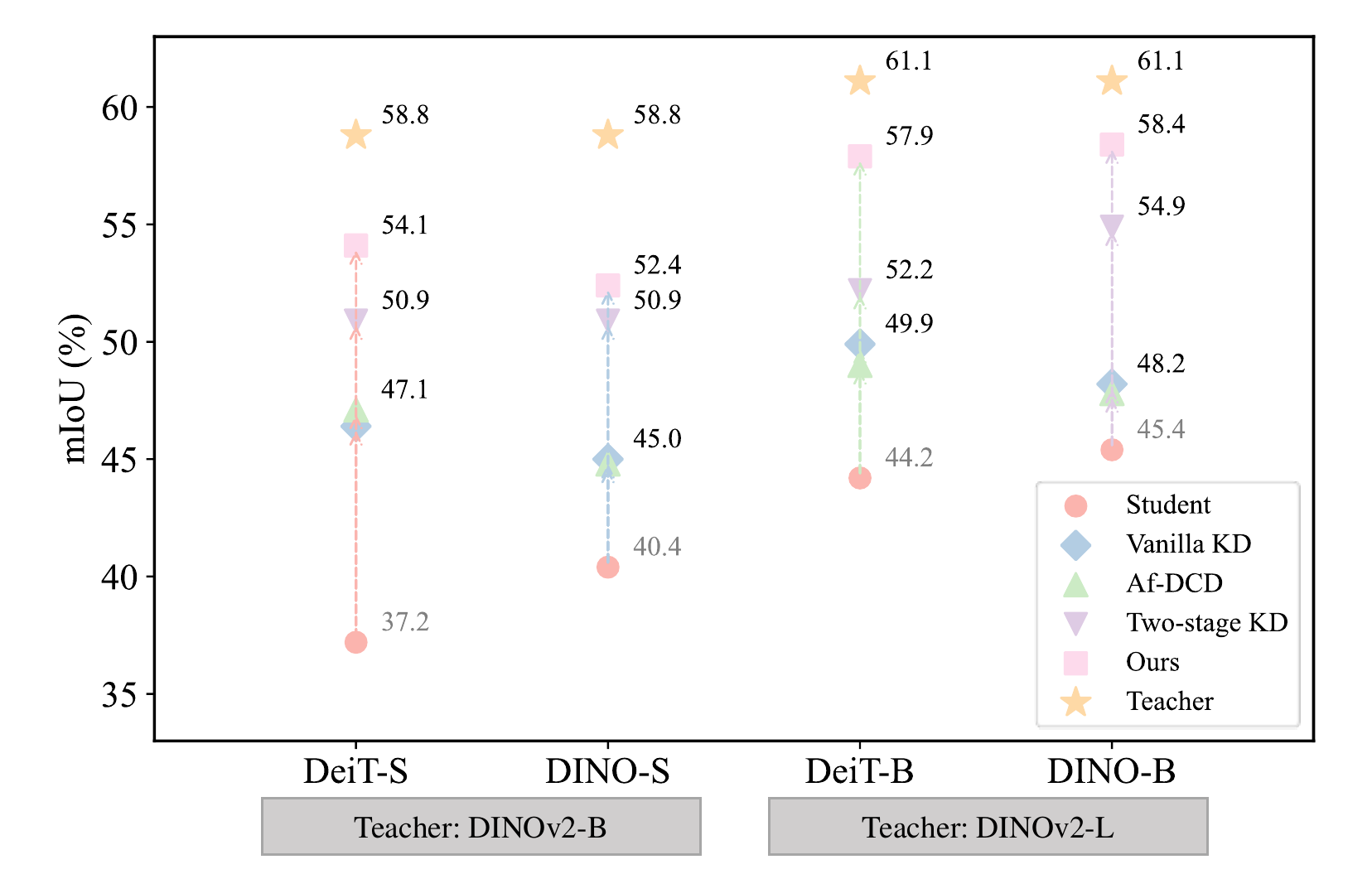}
		\caption{Performance with various KD methods}
		\label{fig:motivation:a}
	\end{subfigure}
	\begin{subfigure}[b]{0.4\linewidth}
		\centering
		\includegraphics[width=\linewidth]{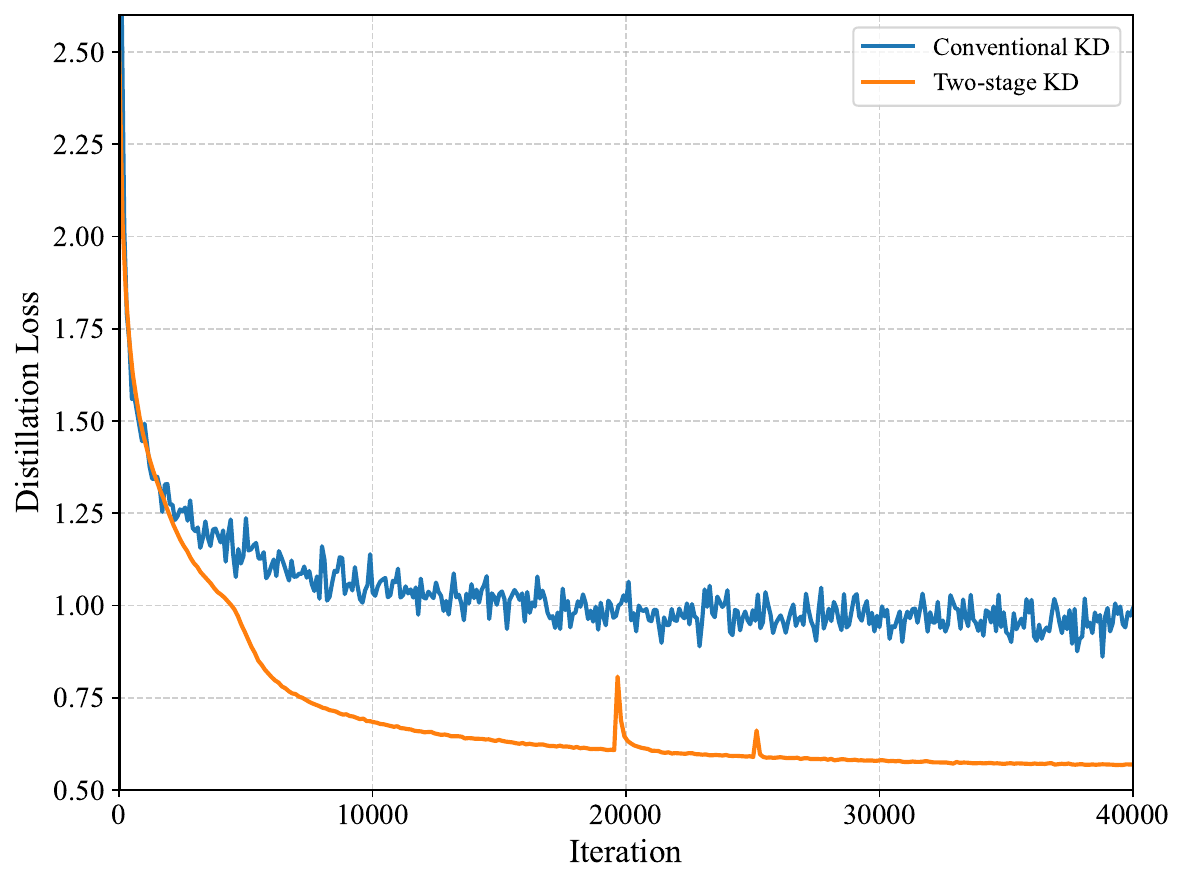}
		\caption{Loss curve}
		\label{fig:motivation:b}
	\end{subfigure}
	\setlength{\abovecaptionskip}{-0 cm}
	\caption{(a) Limited performance gain with conventional KD methods on unseen domains. Two-stage KD effectively improves the generalization performance of student. (b) Loss curves of various KD methods with DINOv2-B $\to$ ViT-S. Conventional single-stage KD causes oscillations and slower convergence, while two-stage KD exhibits smoother loss decay, indicating more stable optimization.}
	\label{fig:motivation}
	\vspace{-0.6cm}
\end{figure*}

\noindent {\bf Conventional Knowledge Distillation.} Knowledge distillation (KD) was initially introduced in~\cite{hinton2015distilling}, where the student learns from hard labels and soft labels obtained from the final layer of the teacher~\cite{zhao2022decoupled,fan2024scalekd}. Most KD methods focus on image classification and fall into logit-based, feature-based or relation-based categories. When extending KD from classification to semantic segmentation, direct feature matching becomes insufficient. Recent KD approaches for semantic segmentation typically aim to transfer structural semantic correlations and inter-class relations from teacher to student~\cite{tian2022adaptive,liu2024bpkd}. IFVD~\cite{wang2020intra} calculates the discrepancy between various class prototypes, compelling the student to replicate the teacher's intra-class affinities. CWD~\cite{shu2021channel} proposes channel-wise distillation to guide the student in mimicking the teacher's semantics along the channel dimension. CIRKD~\cite{yang2022cross} facilitates cross-image distillation at both pixel and region levels to convey structured information. Af-DCD~\cite{fan2023augmentation} introduces a contrastive learning loss to transfer dense, structured local knowledge from teacher to student. While these methods improve in-domain performance, they rarely consider domain generalization, and their effectiveness deteriorates under distribution shift.

\noindent {\bf VFMs Knowledge Distillation.} With the rapid emergence of VFMs, recent studies have investigated how to distill their knowledge into compact models. DeiT~\cite{touvron2021training} introduces a distillation-token strategy enabling data-efficient training of ViTs and achieve competitive classification accuracy. TinyMIM~\cite{ren2023tinymim} systematically explores different distillation recipes for transferring the benefits of large MIM-pretrained ViTs to compact models. SAMI~\cite{xiong2024efficientsam} leverages masked-image pretraining to reconstruct features from SAM~\cite{kirillov2023segment} image encoder for effective visual representation learning. G2SD~\cite{huang2023generic} proposes a generic-to-specific distillation framework to tap the potential of small ViT models under the supervision of large models pre-trained by masked autoencoders. CustomKD~\cite{lee2025customkd} leverages VFMs to enhance the performance of edge models in scenarios with unlabeled data and semi-supervised learning. Proteus~\cite{zhang2025accessing} proposes multi-level distillation objectives to efficiently reproduce the representations of VFMs. Although these methods substantially advance efficiency and task adaptation, they mainly focus on in-domain or task-specific transfer, leaving cross-domain generalization largely unexplored.

\noindent {\bf Discuss.} While some prior works~\cite{huang2023generic,vemulapalli2024knowledge} have explored multi-stage distillation, their designs primarily follow a generic-to-specific paradigm: the student first learns task-agnostic representations, and is then jointly optimized with task supervision and feature or logit distillation to acquire task-specific knowledge. Although this strategy improves in-domain performance, it tends to bias the student toward the source domain, since feature and task objectives are coupled throughout task learning (see \cref{Motivation}). As a result, such designs are inherently task-oriented rather than domain-general. In contrast, we explicitly transfer the out-of-domain robustness of VFMs into compact models. To this end, we adopt a multi-stage schedule where domain-agnostic representation learning is isolated from task optimization, preventing domain overfitting. Furthermore, we introduce a query-based soft distillation mechanism that enables the student to selectively retrieve transferable spatial knowledge from the teacher. Our approach establishes a new paradigm for transferring the generalization ability of VFMs into lightweight models.

%% file: sec/3_methods.tex
\section{Methodology}

\subsection{Preliminary}

\textbf{Domain Generalized Semantic Segmentation} (DGSS) aims to learn domain-invariant representations from labeled source domains and generalize to unseen target domains.
Formally, given labeled sources $D_S=\{(x_S^i, y_S^i)\}_{i=1}^{N_S}$ and unseen targets $D_T=\{x_T^j\}_{j=1}^{N_T}$ (not accessible during training), the segmentation model $\mathcal{F}_{\theta_f}$ is trained by minimizing
\begin{equation}\label{loss_DGSS}
	\begin{aligned}
		\min_{\theta_f}\; \mathbb{E}_{(x_S,y_S)\sim D_S}\big[\mathcal{L}(\mathcal{F}_{\theta_f}(x_S), y_S)\big].
	\end{aligned}
\end{equation} 
The challenge of DGSS lies in ensuring robust generalization on the unseen target domains $D_T$.

\noindent \textbf{Knowledge Distillation} (KD) transfers knowledge from a high-capacity teacher model $\mathcal{F}_{\theta_t}$ to a lightweight student model $\mathcal{F}_{\theta_s}$. The distilled representation $\mathcal{F}_{\theta_t}(x)$ can be defined at different levels, such as intermediate features~\cite{romero2014fitnets,huang2022masked}, logits~\cite{huang2022knowledge,sun2024logit}, or attention maps~\cite{guo2023class,zagoruyko2016paying}. A typical distillation objective is
\begin{equation}\label{loss_distillation}
	\begin{aligned}
	\min_{\theta_s}\; \mathbb{E}_{x \sim D}\big[\|\mathcal{F}_{\theta_t}(x) - \mathcal{F}_{\theta_s}(x)\|_2^2\big],
	\end{aligned}
\end{equation}
where $D$ denotes the training data distribution. Conventional KD mainly improves the student’s performance on the same training distribution $D$, whereas its generalization to unseen domains is rarely considered.

\begin{figure*}[!tbp]
	\centering
	\includegraphics[width=0.99\linewidth]{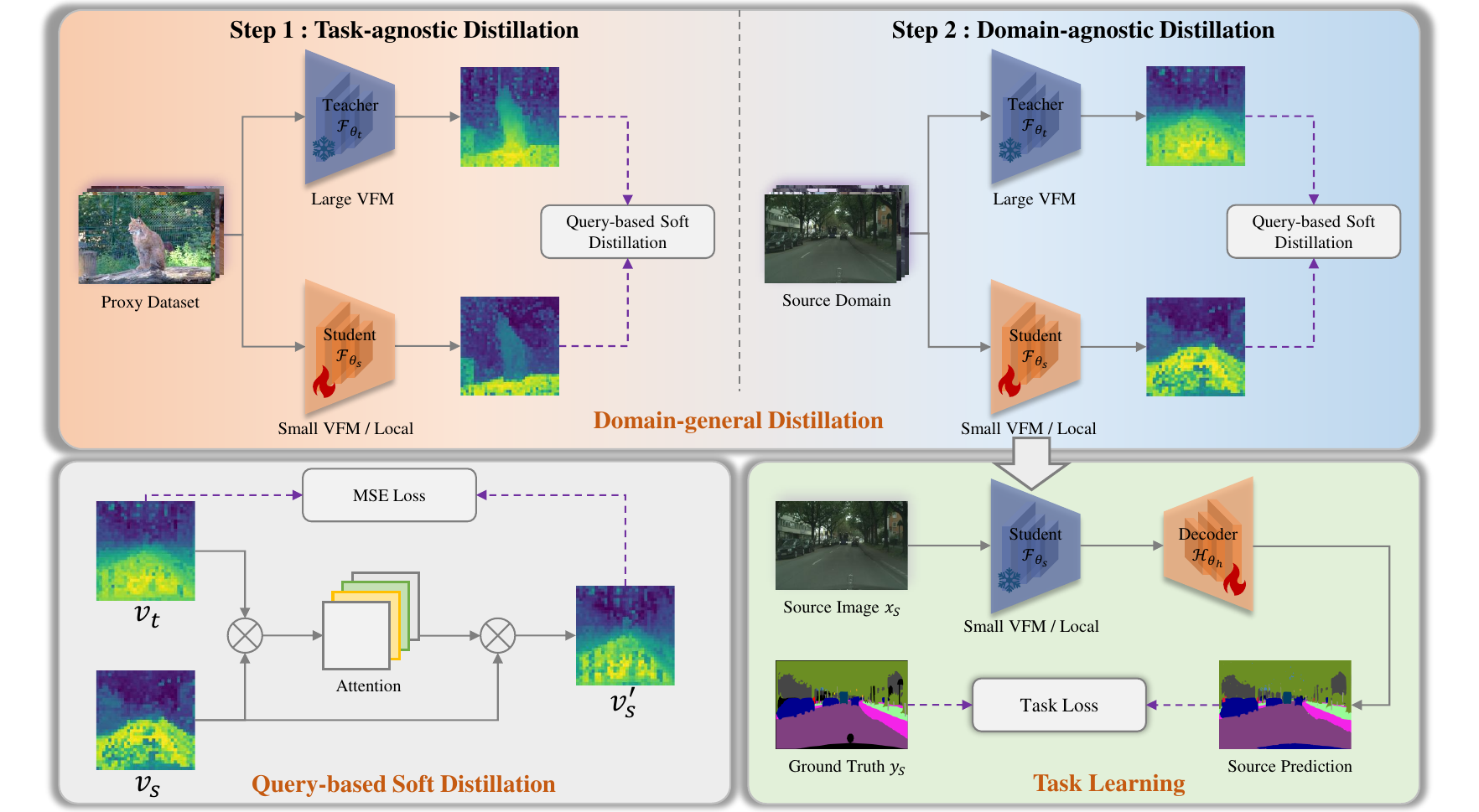}
	\setlength{\abovecaptionskip}{-0 cm}
	\caption{Overview of the proposed GKD framework. GKD comprises two major parts: domain-general distillation and task learning. In the domain-general distillation stage, the student sequentially performs task-agnostic and domain-agnostic distillation, both via the Query-based Soft Distillation mechanism. In the task learning stage, only the decoder is trained on source annotations, while the student encoder is frozen to preserve the domain-general representations.}
	\label{fig:framework}
	\vspace{-0.6cm}
\end{figure*}

\subsection{Motivation Verification}
\label{Motivation}

Before delving into our proposed framework, we conduct preliminary experiments to verify whether transferring generalizable knowledge from VFMs to lightweight students improves the out-of-domain performance. In conventional KD, the student is optimized in a single-stage process where both the task loss and the distillation loss jointly update parameters. Following this setup, the student encoder $\mathcal{F}_{\theta_s}$ can be updated by
\begin{equation}\label{loss_combine}
	\begin{split}
		\min_{\theta_s, \theta_h} \mathbb{E}_{(x_S, y_S) \sim D_S}\big[\mathcal{L}(\mathcal{H}_{\theta_h}(\mathcal{F}_{\theta_s}(x_S)), y_S) \\
		+\|\mathcal{F}_{\theta_t}(x_S) - \mathcal{F}_{\theta_s}(x_S)\|_2^2\big],
	\end{split}
\end{equation}
where $\mathcal{H}$ represents the decoder head parameterized by $\theta_h$. We train the model on the source domain GTAV~\cite{richter2016playing} with various KD methods. We evaluate the generalization performance on unseen target domains Cityscapes~\cite{cordts2016cityscapes}, BDD100K~\cite{yu2020bdd100k}, and Mapillary~\cite{neuhold2017mapillary}. We observe that jointly optimizing feature distillation and task learning tends to hinder the generalization ability of VFMs.

As shown in \cref{fig:motivation:a}, conventional KD yields marginal performance gains over the baseline, while its performance on unseen domains remains notably below the teacher. We attribute this bottleneck to an optimization conflict: the task objective drives the student toward source-specific decision boundaries, while the distillation objective encourages the student to approximate the teacher's domain-invariant representations. These two objectives interfere during training, resulting in unstable convergence and degraded generalization. To verify this hypothesis, we introduce two-stage KD that decouples feature distillation from task learning. Specifically, we first perform feature distillation on source images to enable the student to inherit domain-agnostic representations, and then freeze the encoder to train the decoder with standard task supervision. As illustrated in \cref{fig:motivation:b}, removing the task gradient during representation learning yields more stable optimization and better cross-domain performance. These observations form the foundation of our proposed generalizable KD framework.

\subsection{Proposed Method}

In this section, we propose \textbf{G}eneralizable \textbf{K}nowledge \textbf{D}istillation (\textbf{GKD}), a multi-stage framework that transfers generalizable representations from VFMs to the lightweight student for DGSS, as illustrated in \cref{fig:framework}. GKD consists of two stages: a domain-general distillation stage for representation learning and a task learning stage for downstream segmentation. In the domain-general distillation, the student first distills task-agnostic features from VFMs on a proxy dataset, and then further distills domain-agnostic features from VFMs on the source domains. In task learning, the student encoder is frozen while the decoder is trained with task supervision on labeled source domains, ensuring stable optimization and preserving generalizable representations. To further transfer fine-grained spatial relations, we introduce a Query-based Soft Distillation (QSD) mechanism, which enables the student to retrieve relevant spatial semantics from VFMs and internalize their relational structure.

\noindent {\bf Domain-general Distillation.} VFMs are trained on diverse and massive data~\cite{schuhmann2021laion}, while lightweight student models are typically initialized on ImageNet~\cite{deng2009imagenet}. This creates a representation gap that limits the effectiveness of direct distillation on source data. To mitigate this gap, we split the domain-general stage into two sequential steps.

Inspired by previous work~\cite{zhang2025accessing,vemulapalli2024knowledge}, we first transfer task-agnostic knowledge from VFMs to the student using a proxy dataset $D_P = \{x_P^j\}_{j=1}^{N_P}$ (ImageNet), which is diverse and free of task-specific bias~\cite{torralba2011unbiased,liu2024decade}. This step equips the student with generic visual representations and narrows the initial representation gap. It can be formulated as
\begin{equation}\label{loss_stage_1}
	\begin{aligned}
		\min_{\theta_s} \mathbb{E}_{x_P \sim D_P}\big[\mathcal{L}_{QSD}(\mathcal{F}_{\theta_t}(x_P), \mathcal{F}_{\theta_s}(x_P))\big],
	\end{aligned}
\end{equation}
where $\mathcal{L}_{QSD}$ denotes the proposed query-based soft distillation. Next, the student continues distillation on source images $D_S$, enabling it to encounter task-relevant and domain-agnostic features (e.g., urban objects and scene understanding), without introducing domain-specific supervision bias. Formally, it is defined as
\begin{equation}\label{loss_stage_2}
	\begin{aligned}
		\min_{\theta_s} \mathbb{E}_{x_S \sim D_S}\big[\mathcal{L}_{QSD}(\mathcal{F}_{\theta_t}(x_S), \mathcal{F}_{\theta_s}(x_S))\big].
	\end{aligned}
\end{equation}

\noindent {\bf Task Learning.} After domain-general distillation, we integrate the student encoder with the decoder and optimize with task supervision on labeled source domains
\begin{equation}\label{loss_task_learning}
	\begin{aligned}
		\min_{\theta_h} \mathbb{E}_{(x_S, y_S)\sim D_S}\big[\mathcal{L}(\mathcal{H}_{\theta_h}(\mathcal{F}_{\theta_s}(x_S)), y_S)\big],
	\end{aligned}
\end{equation}
where $\mathcal{L}$ is the segmentation loss (e.g., cross-entropy). This stage ensures that the distilled domain-general representations are effectively grounded into the downstream task.

\begin{figure}[!tbp]
	\centering
	\includegraphics[width=0.99\linewidth]{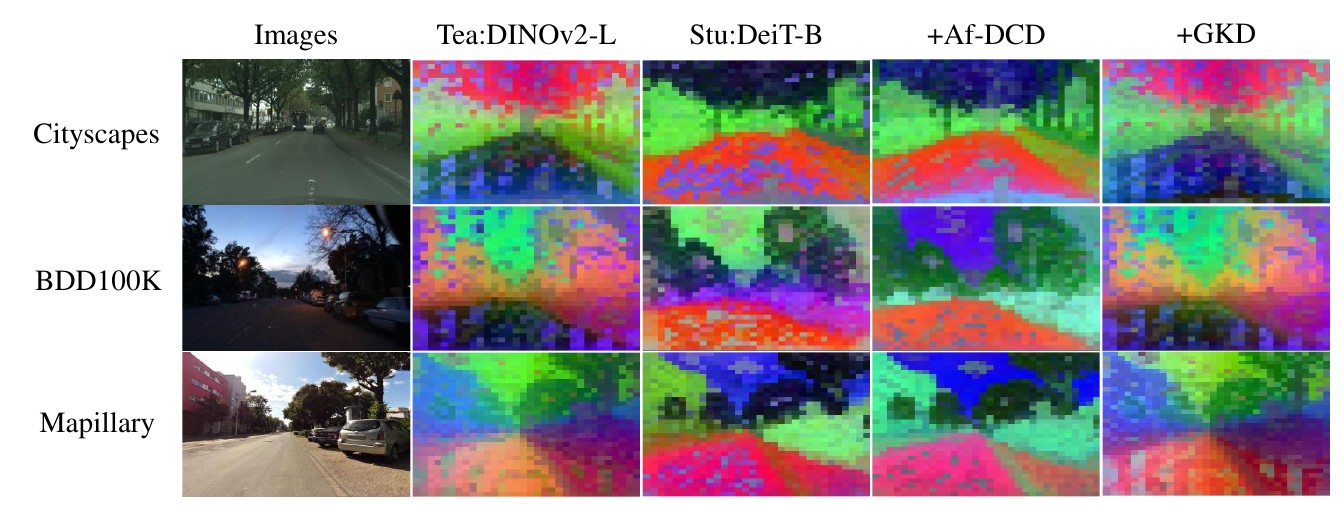}
	\setlength{\abovecaptionskip}{-0 cm}
	\caption{PCA visualization. Feature embedding is extracted from the last layer of encoder. GKD effectively distills the spatial structure information of VFMs.}
	\label{fig:feature}
	\vspace{-0.6cm}
\end{figure}

\noindent {\bf Query-based Soft Distillation.} Existing feature distillation methods typically enforce point-wise alignment between student and teacher features. However, semantic information at corresponding spatial location often differs~\cite{lin2022knowledge,yang2022focal}, point-wise distillation fails to preserve spatial structure and global relational dependencies. As shown in \cref{fig:feature}, VFMs exhibit robust and domain-invariant spatial structure. To transfer these properties, we propose query-based soft distillation (QSD), which enables the student to retrieve all teacher features via attention, and reweights its spatial responses. This allows the student to internalize the teacher's relational structure rather than merely imitate local activations.

Formally, given student features $v_s \in \mathbb{R}^{B \times N \times C_s}$ and teacher features $v_t \in \mathbb{R}^{B \times N \times C_t}$, where $N$ denotes the number of spatial tokens and $C_s$, $C_t$ are the embedding dimensions. To capture the spatial relational dependencies between the teacher and the student, we compute the attention $W \in \mathbb{R}^{B \times N \times N}$
\begin{equation}\label{similarity}
	\begin{aligned}
		& W = \varphi(v_s)\cdot v_t^\top, \\
		& W_{ij} = \langle \varphi(v_s^i), v_t^j \rangle,
	\end{aligned}
\end{equation}
where $\langle \cdot,\cdot \rangle$ represents the inner-product, $\varphi(\cdot)$ is a linear projection layer that adapts $v_s$ to the same dimensions as $v_t$. We then reconstruct the student features based on the attention $W$ as
\begin{equation}\label{reconstructed}
	\begin{aligned}
		v'_s = \sigma(\varphi(v_s)\cdot v_t^\top)\cdot \phi(v_s),
	\end{aligned}
\end{equation}
where $\sigma(\cdot)$ denotes the softmax function, $\phi(\cdot)$ is another linear projection layer. This process redistributes the original student features, enabling each spatial position to integrate intrinsic local information with global context aggregated from teacher features. Finally, we constrain the reconstructed student features to align with teacher features via Mean Squared Error (MSE) loss
\begin{equation}\label{loss_feat}
	\begin{aligned}
		\mathcal{L}_{feat} = \|v'_s - v_t\|_2^2.
	\end{aligned}
\end{equation}

Inspired by DINOv2~\cite{oquab2023dinov2}, we further introduce a masked patch-level distillation objective to reveal the hidden knowledge from VFMs. Specifically, we randomly mask patches in the image and feed the masked image to the student to obtain masked features $v^{mask}_s$. Following the previous procedures, the mask distillation loss is defined as
\begin{equation}\label{loss_mask}
	\begin{aligned}
		\mathcal{L}_{mask} = \|v'^{mask}_s - v_t\|_2^2.
	\end{aligned}
\end{equation}

Additionally, we also perform QSD on CLS token to transfer global semantics. $v_s^{cls}$ and $v_t^{cls}$ denote the CLS token of the student and teacher, respectively. We apply the same reconstruction procedure in \cref{reconstructed} to obtain $v'^{cls}_s$, the CLS distillation loss is defined as
\begin{equation}\label{loss_cls}
	\begin{aligned}
		\mathcal{L}_{cls} = \|v'^{cls}_s - v^{cls}_t\|_2^2.
	\end{aligned}
\end{equation}

The final distillation loss is
\begin{equation}\label{loss}
	\begin{aligned}
		\mathcal{L}_{QSD} = \alpha \mathcal{L}_{feat} + \beta \mathcal{L}_{mask} + \gamma \mathcal{L}_{cls},
	\end{aligned}
\end{equation}
where $\alpha, \beta, \gamma$ are hyperparameters to balance the three terms. we set them to 1 by default in our implementation.

%% file: sec/4_exps.tex
\begin{table*}[!tbp]
	\caption {Performance comparison between proposed GKD and various KD methods in the F2L setting. P-R: Potsdam-RGB, P-I: Potsdam-IRRG, V-I: Vaihingen-IRRG. Tea:Teacher. Stu: student.}
	\centering
	\setlength{\tabcolsep}{4.5pt}
	\resizebox{\textwidth}{!}{
		\begin{tabular}{l|c|c|cccc|ccccc|ccc}
			\hline
			\multirow{2}{*}{Method} & \multirow{2}{*}{Arch} & \multirow{2}{*}{Params} & \multicolumn{4}{c|}{GTAV} & \multicolumn{5}{c|}{Cityscapes} & \multicolumn{3}{c}{P-R} \\
			\cline{4-7} \cline{8-12} \cline{13-15}
			& & & Citys & BDD & Map & Avg. & Night & Snow & Fog & Rain & Avg. & P-I & V-I & Avg. \\
			\hline
			Tea: DINOv2   & ViT-L & 324.8M & 63.3 & 56.1 & 63.9 & 61.1 & 54.6 & 69.4 & 78.9 & 72.6 & 68.9 & 76.7 & 63.4 & 70.1 \\
			\hline
			DINOv2   & ViT-B & 106.8M & 59.6 & 54.3 & 62.6 & 58.8 & 49.9 & 67.6 & 77.5 & 69.9 & 66.2 & 72.3 & 55.9 & 64.1 \\
			\hline
			Stu: DeiT   & ViT-B & 106.8M & 43.1 & 41.8 & 47.7 & 44.2 & 28.1 & 47.2 & 64.1 & 48.5 & 47.0 & 67.5 & 34.1 & 50.8 \\
			+Vanilla KD~\cite{romero2014fitnets}  & ViT-B & 106.8M & 48.5 & 48.2 & 53.2 & 49.9 & 33.2 & 55.7 & 71.3 & 57.0 & 54.3 & 69.0 & 42.9 & 56.0 \\
			+CWD~\cite{shu2021channel}          & ViT-B & 106.8M & 49.3 & 46.7 & 51.8 & 49.3 & 33.1 & 53.8 & 70.5 & 54.1 & 52.9 & 70.0 & 41.5 & 55.8\\
			+Af-DCD~\cite{fan2023augmentation}       & ViT-B & 106.8M & 48.4 & 45.2 & 53.4 & 49.0 & 31.7 & 54.3 & 67.3 & 55.3 & 52.1 & 69.5 & 42.5 & 56.0 \\
            +G2SD~\cite{huang2023generic}        & ViT-B & 106.8M & 50.8 & 49.1 & 53.4 & 51.1 & 33.1 & 55.7 & 69.5 & 56.8 & 53.8 & 72.4 & 46.7 & 59.5 \\
			+Vitkd~\cite{Yang_2024_CVPR}        & ViT-B & 106.8M & 45.1 & 45.6 & 49.8 & 46.8 & 31.1 & 55.2 & 66.3 & 52.8 & 51.4 & 67.8 & 39.8 & 53.8 \\
			+Proteus~\cite{zhang2025accessing} & ViT-B & 106.8M & 48.1 & 46.4 & 52.8 & 49.1 & 32.5 & 54.6 & 69.4 & 54.6 & 52.8 &  70.1 & 43.5 & 56.8 \\
			\rowcolor[gray]{0.85}
			+GKD         & ViT-B & 106.8M & 58.3 & 54.2 & 61.3 & \textbf{57.9} & 43.8 & 69.4 & 76.7 & 68.4 & \textbf{64.6} & 74.5 & 55.6 & \textbf{65.1} \\
			\hline
			Tea: DINOv2   & ViT-B & 106.8M & 59.6 & 54.3 & 62.6 & 58.8 & 49.9 & 67.6 & 77.5 & 69.9 & 66.2 & 72.3 & 55.9 & 64.1 \\ 
			\hline
			DINOv2   & ViT-S & 41.9M & 53.2 & 51.3 & 57.1 & 53.9 & 39.3 & 64.1 & 68.7 & 61.0 & 58.3 & 73.9 & 54.0  & 64.0\\
			\hline
			Stu: DeiT     & ViT-S & 41.9M & 34.9 & 33.8 & 42.8 & 37.2 & 22.7 & 43.0 & 55.0 & 42.2 & 40.7 & 67.6 & 28.7 & 48.2 \\
			+Vanilla KD~\cite{romero2014fitnets}   & ViT-S & 41.9M & 45.0 & 44.2 & 49.9 & 46.4 & 31.4 & 51.3 & 63.6 & 50.1 & 49.1 & 70.3 & 37.6 & 54.0 \\
			+CWD~\cite{shu2021channel}          & ViT-S & 41.9M & 45.9 & 44.5 & 49.8 & 46.7 & 31.7 & 51.0 & 64.7 & 52.6 & 50.0 & 70.4 & 38.6 & 54.5 \\
			+Af-DCD~\cite{fan2023augmentation}       & ViT-S & 41.9M & 44.7 & 45.6 & 50.9 & 47.1 & 31.6 & 49.9 & 70.1 & 49.9 & 50.4 & 71.2 & 38.2 & 54.7 \\
            +G2SD~\cite{huang2023generic}        & ViT-B & 41.9M & 45.2 & 45.9 & 52.3 & 47.8 & 33.5 & 51.4 & 65.6 & 54.2 & 51.2 & 72.7 & 40.2 & 56.5 \\
			+Vitkd~\cite{Yang_2024_CVPR}        & ViT-S & 41.9M & 42.5 & 42.5 & 48.2 & 44.4 & 28.0 & 51.3 & 65.1 & 45.9 & 47.6 & 62.9 & 34.9 & 48.9 \\
			+Proteus~\cite{zhang2025accessing} & ViT-S & 41.9M & 47.4 & 44.6 & 50.2 & 47.4 & 32.8 & 51.3 & 62.1 & 49.3 & 48.9 & 70.8 & 38.5 & 54.7 \\
			\rowcolor[gray]{0.85}
			+GKD         & ViT-S & 41.9M & 54.9 & 49.8 & 57.8 & \textbf{54.1} & 39.3 & 60.4 & 72.7 & 58.4 & \textbf{57.7} & 73.8 & 48.7 & \textbf{61.3} \\
			\hline
		\end{tabular}
	}
	\label{tab:1}
	\vspace{-0.6cm}
\end{table*}

\section{Experiments}
\subsection{Experimental Setup}

\noindent {\bf Datasets.} We evaluate our proposed methods on five driving-scene segmentation datasets that share 19 categories and two cross-urban remote sensing datasets that share 6 categories. In detail, Cityscapes (Citys)~\cite{cordts2016cityscapes} is an autonomous driving dataset that contains 2975 training images and 500 validation images, each with the resolution of 2048$\times$1024. BDD100K (BDD)~\cite{yu2020bdd100k} and Mapillary (Map)~\cite{neuhold2017mapillary} contain 1,000 1280$\times$720 images and 2,000 1920$\times$1080 images for validation, respectively. Adverse Conditions Dataset with Correspondence (ACDC)~\cite{sakaridis2021acdc} is a semantic segmentation dataset that consists of samples from four types of adverse conditions (rain, fog, night and snow). GTAV~\cite{richter2016playing} is a synthetic dataset, which has 24,966 simulated images from the game. ISPRS Potsdam and Vaihingen~\cite{zhang2023pseudo} provide aerial images from two different cities, Potsdam contains 38 images with a size of 6000$\times$6000, and provides both RGB and IR-R-G bands. While Vaihingen has 33 images with a size of 2000$\times$2000 and only IRRG channels. Following existing DGSS methods~\cite{bi2024learning,gong2024crossearth,wei2024stronger}, we employ three evaluation settings: GTAV $\to$ Citys + BDD + Map, Citys $\to$ Night + Snow + Fog + Rain, Potsdam-RGB (P-R) $\to$ Potsdam-IRRG (P-I) + Vaihingen-IRRG (V-I). The evaluation metric is mean Intersection of Union (mIoU).

\noindent {\bf Implementation details.} We use AdamW~\cite{loshchilov2017decoupled} with a learning rate of 5e-4 and a weight decay of 0.05 during distillation. In the F2L setting, all models are first trained on ImageNet for 100 epochs with a batch size of 512, at a resolution of 224$\times$224, and then trained on source domains for 300 epochs with a batch size of 128, at a resolution of 512$\times$512. In the F2F setting, all models are directly trained on source domains for 300 epochs. In task training, the student is integrated with Mask2Former~\cite{cheng2022masked} and inherits the task loss from Mask2Former. We use AdamW with a learning rate of 1e-5 for the backbone and 1e-4 for the decoder. We utilize a configuration of 40,000 iterations with a batch size of 4, and crop images to 512 $\times$ 512.

\begin{table*}[!tbp]
	\caption {Performance comparison between proposed GKD and various KD methods in the F2F setting. TrV: Transform Vision. Tea:Teacher. Stu: student.}
	\centering
	\setlength{\tabcolsep}{4.5pt}
	\resizebox{\textwidth}{!}{
		\begin{tabular}{l|c|c|cccc|ccccc|ccc}
			\hline
			\multirow{2}{*}{Method} & \multirow{2}{*}{Arch} & \multirow{2}{*}{Params} & \multicolumn{4}{c|}{GTAV} & \multicolumn{5}{c|}{Cityscapes} & \multicolumn{3}{c}{P-R} \\
			\cline{4-7} \cline{8-12} \cline{13-15}
			& & & Citys & BDD & Map & Avg. & Night & Snow & Fog & Rain & Avg. & P-I & V-I & Avg. \\
			\hline
			Tea: DINOv2   & ViT-L & 324.8M & 63.3 & 56.1 & 63.9 & 61.1 & 54.6 & 69.4 & 78.9 & 72.6 & 68.9 & 76.7 & 63.4 & 70.1 \\
			\hline
			Stu: DINOv2      & ViT-B & 106.8M & 59.6 & 54.3 & 62.6 & 58.8 & 49.9 & 67.6 & 77.5 & 69.9 & 66.2 & 72.3 & 55.9 & 64.1 \\
			+Vanilla KD~\cite{romero2014fitnets}   & ViT-B & 106.8M & 59.9 & 54.5 & 60.2 & 58.2 & 48.6 & 68.0 & 79.4 & 70.5 & 66.6 & 75.9 & 52.9 & 64.4 \\
			+Af-DCD~\cite{fan2023augmentation}       & ViT-B & 106.8M & 59.5 & 53.0 & 60.0 & 57.5 & 48.9 & 68.0 & 79.1 & 71.0 & 66.7 & 76.2 & 52.3 & 64.3 \\
			+Vitkd\cite{Yang_2024_CVPR}        & ViT-B & 106.8M & 58.0 & 53.0 & 59.3 & 56.7 & 46.6 & 67.6 & 77.1 & 69.6 & 65.2 & 75.5 & 51.6 & 63.6 \\
			+Proteus~\cite{zhang2025accessing} & ViT-B & 106.8M & 60.1 & 54.6 & 61.4 & 58.7 & 48.3 & 67.6 & 79.7 & 71.1 & 66.7 & 75.6 & 53.5 & 64.6 \\
			\rowcolor[gray]{0.85}
			+GKD         & ViT-B & 106.8M & 62.6 & 55.0 & 61.8 & \textbf{59.8} & 48.3 & 71.3 & 80.3 & 72.0 & \textbf{68.0} & 75.4 & 56.4 & \textbf{65.9}   \\
			\hline
			Tea: DINOv2    & ViT-B & 106.8M & 59.6 & 54.3 & 62.6 & 58.8 & 49.9 & 67.6 & 77.5 & 69.9 & 66.2 & 72.3 & 55.9 & 64.1 \\
			\hline
			Stu: DINOv2    & ViT-S & 106.8M & 53.2 & 51.3 & 57.1 & 53.9 & 39.3 & 64.1 & 68.7 & 61.0 & 58.3 & 73.9 & 54.0 & 64.0 \\
			+Vanilla KD~\cite{romero2014fitnets}   & ViT-S & 41.9M & 52.9 & 49.4 & 56.3 & 52.9 & 38.6 & 62.6 & 73.8 & 61.8 & 59.2 & 76.5 & 48.9 & 62.7 \\
			+Af-DCD~\cite{fan2023augmentation}       & ViT-S & 41.9M & 54.0 & 50.2 & 55.7 & 53.3 & 37.5 & 63.2 & 75.4 & 59.3 & 58.8 & 74.2 & 42.6 & 58.4 \\
			+Vitkd~\cite{Yang_2024_CVPR}        & ViT-S & 41.9M & 49.9 & 49.1 & 55.7 & 51.6 & 37.8 & 62.9 & 73.5 & 60.3 & 58.6 & 71.7 & 43.7 & 57.7 \\
			+Proteus~\cite{zhang2025accessing} & ViT-S & 41.9M & 53.5 & 49.7 & 56.9 & 53.4 & 37.6 & 62.5 & 74.9 & 62.6 & 59.4 & 76.0 & 42.5 & 59.3 \\
			\rowcolor[gray]{0.85}
			+GKD         & ViT-S & 41.9M & 57.1 & 51.3 & 58.4 & \textbf{55.6} & 39.2 & 62.6 & 75.5 & 62.2 & \textbf{59.9} & 74.0 & 53.5 & \textbf{63.8}  \\
			\hline
			Tea: EVA02   & TrV-L & 324.8M & 58.4 & 52.5 & 59.0 & 56.7 & 39.1 & 64.9 & 73.3 & 62.6 & 60.0 & 74.8 & 48.8 & 61.8 \\
			\hline
			Stu: EVA02      & TrV-B & 106.8M & 56.2 & 53.0 & 59.4 & 56.2 & 46.1 & 65.1 & 76.7 & 62.6 & 62.6 & 74.7 & 51.6 & 63.2 \\
			+Vanilla KD~\cite{romero2014fitnets}   & TrV-B & 106.8M & 54.4 & 53.2 & 59.4 & 55.7 & 43.5 & 65.2 & 75.3 & 62.5 & 61.6 & 71.4 & 47.4 & 59.4 \\
			+Af-DCD~\cite{fan2023augmentation}       & TrV-B & 106.8M & 55.8 & 52.9 & 58.0 & 55.6 & 46.4 & 65.8 & 75.4 & 63.5 & 62.7 & 72.8 & 47.5 & 60.2 \\
			+Vitkd~\cite{Yang_2024_CVPR}        & TrV-B & 106.8M & 48.6 & 50.2 & 55.2 & 51.3 & 32.5 & 59.2 & 68.4 & 56.6 & 54.2 & 71.2 & 44.2 & 57.7 \\
			+Proteus~\cite{zhang2025accessing} & TrV-B & 106.8M & 53.7 & 52.8 & 59.4 & 55.3 & 45.4 & 64.7 & 74.2 & 61.1 & 61.4 & 73.5 & 48.6 & 61.1 \\
			\rowcolor[gray]{0.85}
			+GKD         & TrV-B & 106.8M & 59.0 & 54.5 & 61.0 & \textbf{58.2} & 46.9 & 67.8 & 77.1 & 65.8 & \textbf{64.4} & 76.4 & 57.9     &\textbf{67.2} \\
			\hline
			Tea: EVA02    & TrV-B & 106.8M & 45.9 & 44.1 & 49.8 & 46.6 & 20.9 & 54.6 & 63.3 & 48.7 & 46.9 & 66.4 & 34.0 & 50.2 \\
			\hline
			Stu: EVA02    & TrV-S & 41.9M & 48.5 & 47.0 & 52.8 & 49.4 & 37.6 & 56.4 & 70.8 & 54.4 & 54.8 & 69.4 & 42.8 & 56.1 \\
			+Vanilla KD~\cite{romero2014fitnets}   & TrV-S & 41.9M & 47.5 & 46.2 & 52.2 & 48.6 & 34.1 & 57.1 & 69.7 & 52.1 & 53.2 & 69.9 & 42.2 & 56.1 \\
			+Af-DCD~\cite{fan2023augmentation}       & TrV-S & 41.9M & 48.3 & 47.2 & 52.1 & 49.2 & 36.1 & 56.8 & 70.9 & 55.0 & 54.7 & 70.3 & 42.6 & 56.5 \\
			+Vitkd~\cite{Yang_2024_CVPR}        & TrV-S & 41.9M & 43.3 & 42.1 & 47.7 & 44.4 & 28.9 & 52.2 & 66.1 & 49.7 & 49.2 & 66.0 & 34.8 & 50.4 \\
			+Proteus~\cite{zhang2025accessing} & TrV-S & 41.9M & 47.1 & 45.1 & 51.1 & 47.8 & 34.0 & 56.3 & 70.8 & 50.6 & 52.9 &  68.4 & 41.8 & 55.1 \\
			\rowcolor[gray]{0.85}
			+GKD         & TrV-S & 41.9M & 51.1 & 45.9 & 53.7 & \textbf{50.2} & 36.0 & 59.0 & 71.2 & 55.8 & \textbf{55.5} & 71.7 & 45.7 & \textbf{58.7} \\
			\hline
		\end{tabular}
	}
	\label{tab:2}
	\vspace{-0.6cm}
\end{table*}

\subsection{Comparison with various KD Methods}

We compare the proposed GKD with conventional KD methods across three different DGSS benchmarks to assess its effectiveness and cross-domain generalization. We adopt two initialization regimes for the student: (1) foundation-to-local (F2L): The locally trained model initialized from ImageNet. (2) foundation-to-foundation (F2F): The small VFMs trained on large-scale datasets.

\begin{figure}[!tbp]
	\centering
	\hspace{-0.3cm}
	\begin{subfigure}[b]{0.49\linewidth}
		\includegraphics[width=\linewidth]{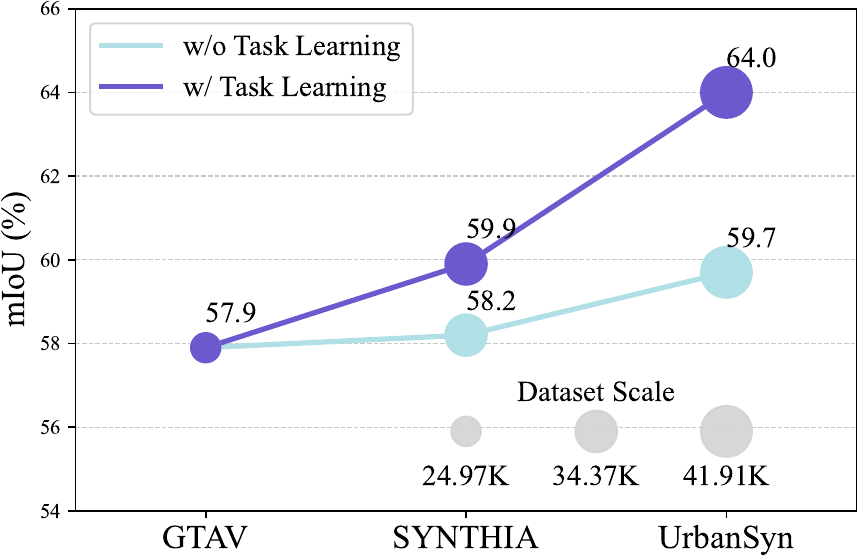}
		\caption{DINOv2-L$\to$ViT-B}
		\label{fig:source_scaling_b}
	\end{subfigure}
	\hspace{-0.1cm}
	\begin{subfigure}[b]{0.49\linewidth}
		\includegraphics[width=\linewidth]{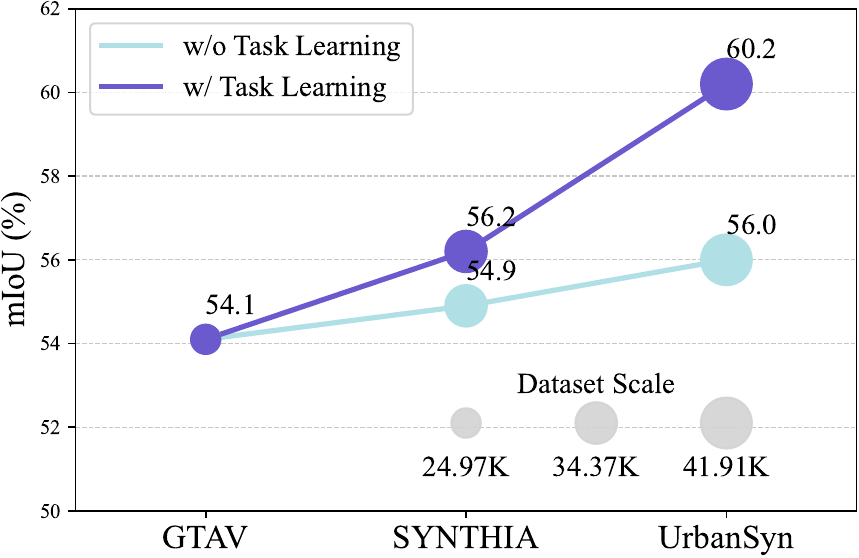}
		\caption{DINOv2-B $\to$ ViT-S}
		\label{fig:source_scaling_s}
	\end{subfigure}
	\setlength{\abovecaptionskip}{-0 cm}
	\caption{Performance comparison on more source domains under \textbf{Citys + BDD + Map} generalization setting. w/o Task Learning: trained with distillation only. w/ Task Learning: trained with both distillation and task training. We use DeiT to initialize the student, (a) and (b) represent different distillation architectures.}
	\label{fig:scaling_up}
	\vspace{-0.6cm}
\end{figure}

\noindent {\bf Results in the F2L setting.} We adopt DeiT~\cite{touvron2021training} as the student and DINOv2 as the teacher. We conduct experiments under stronger settings (DINOv2-L $\to$ ViT-B) and baseline settings (DINOv2-B $\to$ ViT-S), as shown in \cref{tab:1}. GKD significantly outperforms conventional KD methods across all benchmarks. Notably, the official DINOv2-S/B models are distilled from DINOv2-g which has stronger performance, GKD achieve comparable results when trained on ImageNet and source domains. DeiT-B trained with GKD obtains 57.9\% average mIoU on GTAV $\to$ Citys + BDD + Map, close to 58.8\% from DINOv2-B. DeiT-S with GKD even surpasses DINOv2-S by 0.2\%.

\noindent {\bf Results in the F2F setting.} We further evaluate GKD under stronger initialization, where students are initialized from official VFMs (DINOv2~\cite{tian2024learning} and EVA02~\cite{fang2024eva}). As shown in \cref{tab:2}, conventional KD fails to enhance cross-domain generalization, GKD demonstrates consistent and significant improvements. For instance, with DINOv2-B as the student, GKD achieves 59.8\% average mIoU on GTAV $\to$ Citys + BDD + Map, outperforming Vanilla KD by 1.6\%. On the more challenging ACDC target domains, GKD improves the average mIoU from 66.2\% to 68.0\%, and on remote sensing datasets, it raises DINOv2-B from 64.1\% to 65.9\%. Furthermore, GKD transfers effectively to another VFMs EVA02. When applied to EVA02-B, GKD boosts performance by 2.0\% on GTAV $\to$ Citys + BDD + Map and 4.0\% on P-R $\to$ P-I + V-I compared with official checkpoints.

\subsection{Scaling Up}
\begin{table}[!tbp]
	\centering
	\setlength{\tabcolsep}{4.5pt}
	\caption{Performance comparison between the proposed GKD and existing KD methods under different labeled data fractions.}
	\resizebox{\columnwidth}{!}{
		\begin{tabular}{l|cccc|cccc}
			\hline
			\multirow{2}{*}{Method} & \multicolumn{4}{c|}{GTAV} & \multicolumn{4}{c}{Cityscapes} \\
			\cline{2-5} \cline{6-9} 
			& 1/16 & 1/8 & 1/4 & full & 1/16 & 1/8 & 1/4 & full \\
			\hline
			\multicolumn{9}{c}{\textit{F2F}}\\
			\hline
			Stu: DINOv2-B      & 58.3 & 58.4 & 58.7 & 58.8 & 62.1 & 63.9 & 64.1 & 66.2 \\
			+Af-DCD       & 56.2 & 56.6 & 56.9 & 57.5 & 60.9 & 62.2 & 63.9 & 66.7 \\
			\rowcolor[gray]{0.85}
			+GKD         & \textbf{58.4} & \textbf{58.6} & \textbf{59.1} & \textbf{59.8} & \textbf{62.3} & \textbf{64.7} & \textbf{64.9} & \textbf{67.2} \\
			\hline
			Stu: DINOv2-S      & 52.4 & 53.0 & 53.5 & 53.9 & 53.0 & 55.0 & 56.8 & 58.3 \\
			+Af-DCD       & 48.9 & 49.7 & 52.4 & 53.3 & 52.3 & 55.2 & 57.4 & 58.8 \\
			\rowcolor[gray]{0.85}
			+GKD         & \textbf{54.7} & \textbf{55.0} & \textbf{55.3} & \textbf{55.6} & \textbf{54.7} & \textbf{56.5} & \textbf{57.6} & \textbf{59.9} \\
			\hline
			\multicolumn{9}{c}{\textit{F2L}}\\
			\hline
			Stu: DeiT-B      & 42.1 & 42.1 & 43.2 & 44.2 & 39.1 & 42.6 & 43.4 & 47.0 \\
			+Af-DCD       & 47.4 & 48.4 & 49.0 & 49.0 & 45.2 & 48.9 & 50.4 & 52.1 \\
			\rowcolor[gray]{0.85}
			+GKD         & \textbf{56.5} & \textbf{56.8} & \textbf{56.8} & \textbf{57.9} & \textbf{59.1} & \textbf{60.0} & \textbf{61.2} & \textbf{64.6} \\
			\hline
			Stu: DeiT-S      & 35.7 & 36.7 & 37.5 & 37.2 & 32.7 & 38.0 & 38.2 & 40.7 \\
			+Af-DCD       & 46.0 & 46.1 & 46.2 & 47.1 & 43.6 & 46.5 & 49.0 & 50.4 \\
			\rowcolor[gray]{0.85}
			+GKD         & \textbf{51.4} & \textbf{51.5} & \textbf{53.6} & \textbf{54.1} & \textbf{54.6} & \textbf{54.8} & \textbf{57.0} & \textbf{57.7} \\
			\hline
		\end{tabular}
	}
	\label{tab:3}
	\vspace{-0.6cm}
\end{table}

\noindent {\bf Generalization on more Source Domains.} To investigate the effect of multiple source domains, we progressively augment GTAV with two synthetic datasets, SYNTHIA~\cite{ros2016synthia} and UrbanSyn~\cite{gomez2025all}. We evaluate two configurations: (1) SYNTHIA and UrbanSyn are used solely for distillation, while task training relies exclusively on GTAV; and (2) SYNTHIA and UrbanSyn are also included in task training. As shown in \cref{fig:scaling_up}, the performance of student steadily improves as more source domains are incorporated. Notably, even when SYNTHIA and UrbanSyn are used only for distillation, GKD still benefits from the richer visual representations learned from multiple source domains, confirming that GKD effectively transfers domain-agnostic knowledge diverse visual distributions.

\noindent {\bf Generalization on Limited Labeled Data.} We evaluate GKD in label-scarce scenarios by reducing the annotated data to 1/16, 1/8, and 1/4 of the full dataset, as shown in \cref{tab:3}. Leveraging the multi-stage distillation mechanism, the student acquires rich semantic representations, reducing reliance on task annotations. Consequently, even with limited labels, the student maintains strong generalization. In the F2L setting, DeiT-S trained with GKD achieves 51.4\% mIoU on Citys + BDD + Map with only 1/16 labels, outperforming Af-DCD by 5.4\% and the vanilla student by 15.7\%. Similar improvements are observed across other label fractions and target domains, with notable gains for locally trained models. In the F2F setting, GKD continues to demonstrate consistent improvements, highlighting its robustness across various student capacities.

\begin{figure}[!t]
	\centering
	\begin{subfigure}[b]{0.49\linewidth}
		\includegraphics[width=\linewidth]{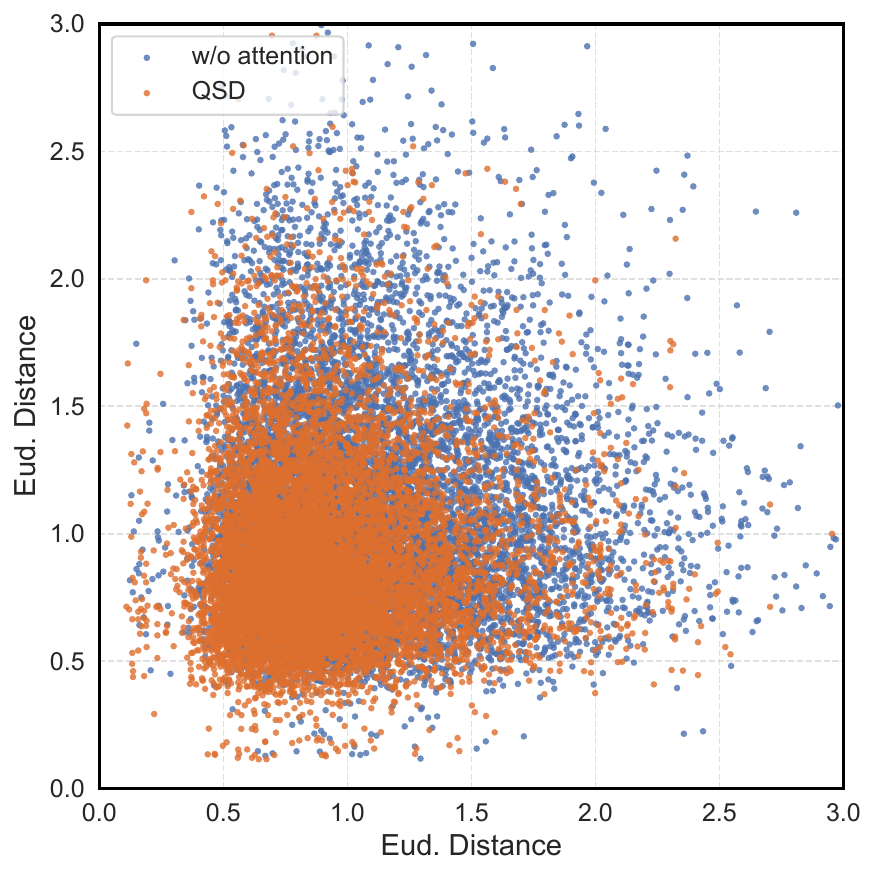}
		\caption{Feature distance}
		\label{fig:scatter_query}
	\end{subfigure}
	\begin{subfigure}[b]{0.49\linewidth}
		\includegraphics[width=\linewidth]{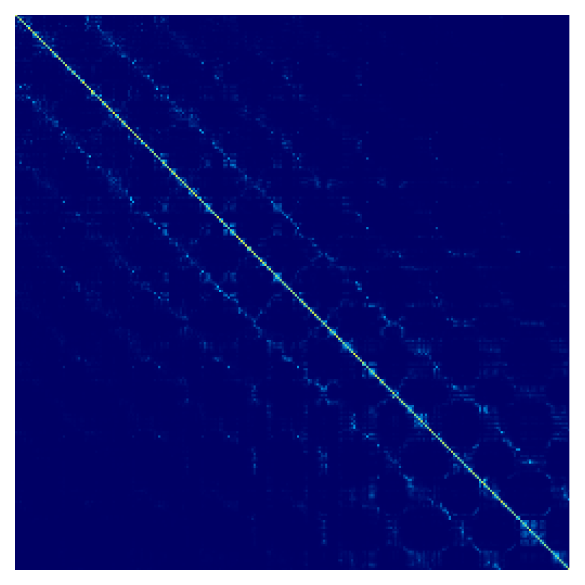}
		\caption{Attention}
		\label{fig:atten_vis}
	\end{subfigure}
	\setlength{\abovecaptionskip}{-0 cm}
	\caption{Visualization of feature distance and attention map. Feature embedding is extracted from the last layer of encoder. We obtain fine-grained representations from the target domains (Citys+BDD+Map) in F2L setting. In (a), we randomly select 10K fine-grained representations measure the distance between the student and the teacher.}
	\vspace{-0.3cm}
	\label{fig:vis}
\end{figure}

\begin{table}[!tbp]
	\centering
	\caption {Ablation study on distillation strategies with DINOv2-B $\to$ ViT-S under GTAV $\to$ Citys + BDD + Map generalization setting. $^{\dagger}$ denotes one-stage KD without decoupled optimization.}
	\begin{tabular}{lcccc}
		\toprule
		Methods & Citys & BDD & Map & Avg. \\
		\midrule
		MSE$^{\dagger}$        &  45.0 & 44.2 & 49.9 & 46.4    \\
        QSD$^{\dagger}$        &  48.9 & 46.5 & 51.1 & 48.8                               \\
		MSE                    &  54.2 & 49.0 & 56.1 & 53.1    \\
        CWD                    &  53.0 & 48.9 & 53.8 & 51.9    \\
		Vitkd                  &  53.2 & 48.7 & 55.0 & 52.3    \\
		QSD                    &  \textbf{54.9} & \textbf{49.8} & \textbf{57.8} & \textbf{54.1}    \\
		\bottomrule
	\end{tabular}
	\label{tab:6}
	\vspace{-0.6cm}
\end{table}

\subsection{Visualization}

We visualize the feature distances and the associated attention to understand the role of proposed Query-based Soft Distillation (QSD). As shown in \cref{fig:scatter_query}, student features trained with QSD exhibit smaller and more compact Euclidean distances to the teacher, indicating better feature alignment. Meanwhile, the attention in \cref{fig:atten_vis} reveals a strong diagonal pattern, indicating that QSD maintains spatial correspondence between the student and teacher. The off-diagonal responses show that the student also selectively aggregates semantics from related teacher features. This selective aggregation enables the student to internalize the teacher’s domain-invariant structure, rather than merely imitating local activations, which is crucial for robust cross-domain generalization.

\begin{table}[!t]
	\caption {Ablation study for each component with DINOv2-B $\to$ ViT-S under GTAV $\to$ Citys + BDD + Map generalization setting.}
	\centering
	\setlength{\tabcolsep}{4pt}
	\resizebox{\columnwidth}{!}{
		\begin{tabular}{ccccccc}
			\toprule
			\multirow{2}{*}{\begin{tabular}[c]{@{}c@{}}Task-agnostic\\ Distillation\end{tabular}} & \multirow{2}{*}{\begin{tabular}[c]{@{}c@{}}Domain-agnostic\\ Distillation\end{tabular}} &
			\multicolumn{3}{c}{QSD}          &
			\multirow{2}{*}{\begin{tabular}[c]{@{}c@{}}Frozen\\ Encoder\end{tabular}} &  \multirow{2}{*}{mIoU} \\ \cline{3-5}
			&                                                                                         & CLS Token & Feature & Mask Patch & &                       \\
			\midrule
			\XSolidBrush           & \XSolidBrush             & \XSolidBrush    & \XSolidBrush & \XSolidBrush & \XSolidBrush &   46.4         \\
			\XSolidBrush           & \Checkmark             & \XSolidBrush      & \XSolidBrush & \XSolidBrush & \XSolidBrush & 50.9         \\
			\Checkmark           & \Checkmark             & \XSolidBrush        & \XSolidBrush & \XSolidBrush & \XSolidBrush &    53.1         \\
            \Checkmark           & \Checkmark             & \Checkmark        & \Checkmark & \XSolidBrush & \XSolidBrush &    53.4         \\
			\Checkmark           & \Checkmark             & \Checkmark        & \Checkmark & \Checkmark & \XSolidBrush &    54.0         \\
			\Checkmark           & \Checkmark             & \Checkmark          & \Checkmark & \Checkmark & \Checkmark &   \textbf{54.1}        \\
			\bottomrule
		\end{tabular}
	}
	\label{tab:7}
	\vspace{-0.6cm}
\end{table}

\subsection{Ablation Study and Analysis}

\noindent {\bf Distillation Strategies.} As shown in \cref{tab:6}, the single-stage KD variants MSE$^{\dagger}$ and QSD$^{\dagger}$ perform significantly worse. In contrast, conventional KD and its enhanced variants under multi-stage schedule achieve competitive results, while the proposed QSD further enhances cross-domain generalization. These results confirm that multi-stage optimization and the relational spatial knowledge are crucial for effectively transferring domain-general representations.

\noindent {\bf Ablation Study.} In \cref{tab:7}, we analyze the contribution of the proposed components. Domain-agnostic distillation contributes most of the gain, and task-agnostic distillation further improves performance. Within QSD, enabling all three distillation objectives yields the best result. Freezing the encoder during task learning prevents domain-general representations from being biased towards the source domain and gives a small gain, while reducing training cost.

%% file: sec/5_conclusion.tex
\section{Conclusion}
Conventional KD preserves accuracy within the same domain but overlooks generalization to unseen domains. In this paper, we present a generalizable knowledge distillation framework that transfers the robust generalization ability of VFMs to compact models. By decoupling domain-agnostic representation learning from task-specific adaptation and integrating a Query-based Soft Distillation (QSD) mechanism, GKD selectively transfers transferable spatial knowledge while mitigating domain overfitting. Extensive experiments across diverse domain generalization benchmarks demonstrate that GKD consistently outperforms existing KD methods under both F2L and F2F settings, achieves strong performance with limited annotations, and scales effectively with additional source domains.